\documentclass{article}

\usepackage{PRIMEarxiv}

\usepackage[utf8]{inputenc} 
\usepackage[T1]{fontenc}    
\usepackage{setspace}
\usepackage{hyperref}       
\usepackage{url}            
\usepackage{booktabs}       
\usepackage{amsfonts}       
\usepackage{nicefrac}       
\usepackage{microtype}      
\usepackage{fancyhdr}       
\usepackage{graphicx}       
\graphicspath{{media/}}     

\usepackage{cite}
\usepackage{amsmath,amssymb,upgreek}
\usepackage[ruled,vlined,linesnumbered]{algorithm2e}
\usepackage[inline]{enumitem}
\DeclareMathOperator*{\argmin}{arg\,min}
\SetAlFnt{\small}
\SetAlCapFnt{\small}
\SetAlCapNameFnt{\small}
\SetKwComment{Comment}{ $\triangleright$\ }{}%
\SetCommentSty{footnotesize}
\SetKw{KwEach}{each}
\SetKw{KwIn}{Input}
\SetKw{KwOut}{Output}
\SetKw{Kwin}{in}
\SetKw{KwBreak}{break}
\usepackage{algorithmic}
\algsetup{linenosize=\tiny}

\usepackage{textcomp}
\usepackage{xcolor}
\usepackage{multirow}
\usepackage[flushleft]{threeparttable}
\usepackage[style=default]{caption}
\def\BibTeX{{\rm B\kern-.05em{\sc i\kern-.025em b}\kern-.08em
    T\kern-.1667em\lower.7ex\hbox{E}\kern-.125emX}}

\title{Multi-surrogate Assisted Efficient Global Optimization for Discrete Problems
}

\author{
  Qi Huang\thanks{The corresponding author}, Roy de Winter, Bas van Stein, Thomas B\"{a}ck, Anna V. Kononova \\
  Leiden Institute of Advanced Computer Science \\
  Leiden University \\
  Leiden, Netherlands\\
  \texttt{\{q.huang, r.de.winter, b.van.stein, t.h.w.baeck, a.kononova\}@liacs.leidenuniv.nl} \\
}

\begin{document}
\maketitle

\begin{abstract}
Decades of progress in simulation-based surrogate-assisted optimization and unprecedented growth in computational power have enabled researchers and practitioners to optimize previously intractable complex engineering problems. This paper investigates the possible benefit of a concurrent utilization of multiple simulation-based surrogate models to solve complex discrete optimization problems. To fulfill this, the so-called Self-Adaptive Multi-surrogate Assisted Efficient Global Optimization algorithm (SAMA-DiEGO), which features a two-stage online model management strategy, is proposed and further bench\-marked on fifteen binary-encoded combinatorial and fifteen ordinal problems against several state-of-the-art non-surrogate or single surrogate assisted optimization algorithms. Our findings indicate that SAMA-DiEGO can rapidly converge to better solutions on a majority of the test problems which shows the feasibility and advantage of using multiple surrogate models in optimizing discrete problems.
\end{abstract}

\keywords{Discrete Optimization \and Surrogate Assisted Optimization \and Bayesian Optimization \and Global Optimization}

\section{Introduction}
Due to the increase in computational power, it becomes feasible in many engineering fields to optimize new or existing components to ever-increasing demands in terms of sustainability, efficiency, and safety. These challenges and arising expectations to tackle them, give space to the development and application of mathematical optimization, which comprises a series of techniques and can be found nearly everywhere in modern-day engineering fields, e.g., aircraft/vehicle design~\cite{giannakoglou_aerodynamic_2006} and chemical engineering~\cite{korovina_chembo_2019}. Nowadays, instead of manually finding the best solutions or designs, a computer with its enormous computational power can tirelessly and smartly go through the search space of these models under certain guidelines called optimization algorithms. 

The primary goal of mathematical optimization is to find a desirable (optimal) solution(s) from a set of candidate solutions based on their performance on objective (target) function(s). Assuming the outputs of objective functions are accessible within reasonable computation time, a group of well-established solvers is the query-based solvers, e.g., evolutionary algorithms and the covariance matrix adaptation evolution strategies. However, in real-world applications, there are cases where the objective functions are expensive to evaluate. Consequently, it is infeasible to allow a solver to evaluate thousands of (feasible) solutions using expensive objective functions due to the unacceptable need for execution time and resources. To tackle expensive optimization problems, Jones et al.~\cite{jones_efficient_1998} proposed the well-known Efficient Global Optimization (EGO), a.k.a, Bayesian Optimization (BO) algorithm, which efficiently solves expensive optimization problems by constructing a so-called surrogate model (a.k.a., meta-model or meta-heuristic) to simulate the objective function and shifting the optimization process that was once based on an expensive objective function to a cheaper surrogate model.

Decades of development have resulted in a wide variety of surrogate-assisted algorithms, each with distinct characteristics and suitable application scenarios. Bhosekar and Ierapetritou~\cite{bhosekar_advances_2018} made an introductory review in 2018, which summarized frequently used surrogate models; famous derivative-free optimization algorithms (solvers) that are suitable to combine with surrogates; as well as accompanied sampling strategies and validation metrics for surrogate models. 
However, the majority of existing surrogate-assisted/based optimization reviews and works are primarily focused on handling (expensive) optimization problems defined in continuous search spaces. This type of search space is exclusively comprised of numerical variables, each of which has an infinite number of values between any two values. Obviously, there exist real-world optimization problems, e.g., vehicle routing and scheduling~\cite{sbihi_combinatorial_2007}, that is entirely built upon a discrete search space, where each variable has a finite number of choices.

In contrast to the flourished development of EGO as well as online surrogate management for real-valued optimization, Jin~\cite{jin_surrogate-assisted_2011} pointed out the scarcity of studies on surrogate-based/assisted discrete/combinatorial optimization in 2011. Later in 2017, a survey by Bartz-Beielstein and Zaefferer~\cite{bartz-beielstein_model-based_2017} identified the key issue as finding suitable surrogate models for discrete problems. This challenge becomes critical if EGO algorithms are applied, as these solvers normally rely on a single surrogate model predefined beforehand.
Therefore, it is a natural choice to consider the possibility of concurrently using multiple surrogate models, which are all capable of handling discrete parameter spaces, to tackle complex discrete problems. Hence, this research focuses on the feasibility and potential benefit of utilizing multiple surrogate models in an EGO-styled algorithm to solve purely discrete problems. To the best of our knowledge, there is a lack of noteworthy studies on this topic.

Given this motivation, a novel multiple-surrogate-assisted optimization algorithm for discrete optimization problems is proposed in this manuscript. The performance of the algorithm is evaluated on both benchmark combinatorial, and integer-valued problems. The algorithm is based on the well-established Efficient Global Optimization (EGO)~\cite{jones_efficient_1998} and features a two-stage online model management strategy inspired by~\cite{bagheri_online_2016}. To address the term \textit{multiple}, four types and a total of thirty-one surrogate models that universally accommodate discrete problems are implemented. Additionally, the source codes and supplementary materials can be found in~\cite{qi_huang_2022_6684443}.

The remainder of the paper is organized as follows: In Section~\ref{sec:related_work}, the relevant literature on online management of multiple surrogate models and feasible surrogate models for discrete optimization are provided. The proposed multi-surrogate-assisted algorithm is presented and block-by-block introduced in Section~\ref{sec:alg}. Next, the algorithms are benchmarked with the settings outlined in Section~\ref{sec:exp setup}, and the experimental results are discussed in Section~\ref{sec:results}. Lastly, the conclusions and suggestions for future work is given in Section~\ref{sec:conclusion}.

\section{Related Literature}
\label{sec:related_work}
\subsection{Online Management of Multiple Surrogate Models}
\label{sec:related_work_A}
Utilizing multiple surrogate models in optimization is a well-practiced idea. Online management of multiple surrogate models denotes the process of maintaining and determining the most suitable proxy (surrogate model) for an objective function while doing optimization. In contrast to offline model management, online management is capable of utilizing incrementally obtained new data samples.
Gorissen et al.~\cite{gorissen_evolutionary_2009} proposed an Evolutionary Model Selection (EMS) algorithm to determine the best surrogate model type and its hyperparameter by minimizing  cross-validation error or Akaike Information Criterion using a modified genetic algorithm. 
Bagheri et al.~\cite{bagheri_online_2016} selected the best-performed type of radial basis function interpolation on the basis of their median absolute errors obtained on newly acquired data samples (unseen to surrogate models). 


Instead of only using the best surrogate out of all candidates, an alter\-native choice is to create the ensemble (aggregation) of multiple surrogate models through computing a weighted sum of predictions produced by all candidate mod\-els trained on existing observations~\cite{hanse_optimally_2022} or on heterogeneous features extracted/selected from data samples~\cite{guo_heterogeneous_2019}.
Further, it is feasible to adaptively merge a group of surrogate models with regard to fidelity level, i.e., reliability or accuracy of models in approximating the original problem~\cite{forrester_multi-fidelity_2007}. 

By all means, the previously described methods aim to select/create a surrogate model out of all candidate models. An alternative strategy is proposed in~\cite{viana_efficient_2013}, namely, the Multiple Surrogate Efficient Global Optimization (MSEGO) algorithm. The MSEGO samples new infill points by independently maximizing the expected improvement on all of its surrogate models, and later evaluates all the new infill points samples using the real function. A similar idea is re-visited in~\cite{wang_comparative_2016} and further developed in~\cite{beaucaire_multi-point_2019}. In comparison to exclusively using EI in MSEGO, Beaucair et al.~\cite{beaucaire_multi-point_2019} suggest employing different infill criteria for different types of surrogate models and therefore concurrently sampling new points from these independent infill criteria. Besides EGO, studies on surrogate-assisted evolutionary algorithms also feature concurrent utilization of multiple surrogates, e.g., Li et al~\cite{li_ensemble_2019} simultaneously used a radial basis function model and a polynomial regression model to assist particle swarm optimization algorithms to converge on expensive continuous problems in the late stage of optimization.

\subsection{Surrogate Modeling for Discrete Problems}
The need for adapting EGO for discrete problems has recently and constantly been emphasized in solving expensive mixed-integer optimization problems. A notable example is hyper\-parameter tuning and automatic model selection for machine learning tasks, where, the EGO-styled solvers are more commonly referred to as Sequential Model-Based Optimization (SMBO) algorithms~\cite{hutter_sequential_2011}. Studying the surrogate models used in SMBO algorithms is a solid entry point for studying the state-of-the-art methodology of applying surrogate models to expensive discrete problems. The application scenario is: given a set of mixed-integer configurations as $\lambda$ and the corresponding performance of the expensive machine learning model as $y$, surrogate models in SMBO are commonly required to model the conditional probability $p(y|\lambda)$. With respect to the types of the surrogate model applied in the well-established SMBO algorithms: Sequential Model-based Algorithm Configuration (SMAC)~\cite{hutter_sequential_2011} and Mixed-Integer Parallel Efficient Global Optimization (MIP-EGO)~\cite{stein_automatic_2019} use random forest regression; Gaussian process regression (a.k.a Kriging) are utilized in Spearmint optimizer~\cite{snoek_practical_2012} and Gaussian Process optimizer (GP optimizer)~\cite{bergstra_algorithms_2011}; adaptive Tree-structured Parzen Estimator (TPE)~\cite{bergstra_algorithms_2011}, instead of modeling the $p(y|\lambda)$, it applies a tree-structure Parzen estimator to concurrently model $p(y|\lambda\geq a)$ and $p(y|\lambda < a)$, where $a$ is a moving threshold; lastly, Bayesian Optimization and HyperBand (BOHB)~\cite{falkner_bohb_2018} resembles multiple TPE models as its back-end surrogate. Other eye-catching examples are: Zaefferer et al.~\cite{zaefferer_efficient_2014} demonstrated the feasibility of using Radial Basis Function networks (RBFn) to solve combinatorial optimization problems; Support Vector Regression (SVR) and gradient boosting regression tree are applied in~\cite{dushatskiy_novel_2021} to model machine learning tasks. To come to the point, all the mentioned surrogate models can naturally handle or be modified to handle discrete variables (e.g. binary, ordinal, and categorical) or structures (tree, graph, etc).

\section{SAMA-DiEGO}
\label{sec:alg}
A novel algorithm, namely, the Self-Adaptive Multi-surrogate Assisted Discrete Efficient Optimization Algorithm (SAMA-DiEGO) is discussed in this section. Firstly, the formulation of the optimization task is given. The algorithm is proposed to solve a single objective optimization problem. For minimization problems\footnote{A maximization problem can be transformed into a minimization problem by multiplying the objective value with -1.}, the optimization task $f\colon \mathcal{S} \rightarrow \mathbb{R}$, can be defined as follows:
\begin{equation}
\label{eq: optimization}
\mathbf{X} = \{x\ |\ \argmin _{x\in \mathcal{S}} f(x),\ f(x)\in \mathbb{R}\}, 
\end{equation}
where $\mathcal{S}$ is the search space that contains all feasible solutions and $\mathbf{X}$ are the optimal solution(s). On that basis, a competent data-driven surrogate model $\hat{f}$ to $f\colon \mathcal{S} \rightarrow \mathbb{R}$ is expected to bear the following properties,
\begin{equation}
\begin{gathered}
\label{eq: surrogate}
\hat{F} = \{ f(x)\ |\ x \in \hat{\mathcal{S}},\  \hat{\mathcal{S}} \subseteq \mathcal{S}\},\\
\hat{f}_{\hat{\mathcal{S}},\hat{F}}(x)\colon \mathcal{S} \rightarrow \mathbb{R},\\
\exists \mathcal{S^*} \subseteq \mathbf{S}\colon \lvert S^* \rvert > 1, 
\forall \hat{x} \in \mathcal{S^*} \colon \hat{f}_{\hat{\mathcal{S}},\hat{F}}(x) \simeq f(x),
\end{gathered}
\end{equation}
where $\hat{\mathcal{S}}$ is the set of available observable input from the solution space $\mathcal{S}$ and their corresponding real objective values of $\hat{\mathcal{S}}$ are recorded in $\hat{F}$. The surrogate model $\hat{f}_{\hat{\mathcal{S}}, \hat{F}}$ learns (e.g. curve fitting) the data ($\hat{\mathcal{S}}$ and $\hat{F}$) and therefore, produces simulated objective values of $f$ on some solutions ($\mathcal{S^*}$) from $\mathcal{S}$. If the surrogate model can accurately simulate the original functions, i.e., given the same inputs, the surrogate model can generate similarly or even the same outputs as the original functions using less computational resources, it is possible to efficiently locate the global optimum of the original functions through exclusively doing optimization using the surrogate as a computationally cheap substitute objective.

Following the notations of the problem formulation, a pseudocode of the SAMA-DiEGO algorithm is presented in Alg.~\ref{alg:SAMA-DiEGO}. The algorithm consists of two stages in general, namely an initialization stage and an optimization stage. The initialization stage determines the initial data sample and based on the initial samples to select the pool of candidate surrogate models for the next stage. The optimization stage iteratively decides the best surrogate model, searches for a promising solution on the determined model, and evaluates the solution using the real objective function. Before diving into details, explanations of three important choices made for SAMA-DiEGO are first discussed. These choices distinguish SAMA-DiEGO from the prototype~\cite{jones_efficient_1998}, and other EGO algorithms (e.g. SMAC~\cite{hutter_sequential_2011}).

\noindent \textbf{Infill Criterion:} Generally speaking, most of the Bayesian optimization rely on a probabilistic-based infill criterion (a.k.a. acquisition function) to find promising points~\cite{wang_new_2017}. One of the well-known and commonly used infill criteria is the Expected Improvement (EI). Assume given a surrogate model $g(x)$ computed on a minimization problem and a reasonable assumption that the model follows a normal distribution, the definition of EI can be written as follows:
\begin{equation}
\begin{aligned}
EI_{g}(\mathbf{x})=&(g^{*}-g(x)) \Phi\left(\frac{g^{* }-g(x)}{\sigma(g(x))}\right)\\
&+\sigma(g(x)) \phi\left(\frac{g^{* }-g(x)}{\sigma(g(x))}\right),
\end{aligned}
\end{equation}
where $g^{*}$ is the best-so-far found objective value, $g(x)$ and $\sigma(g(x))$ are the prediction and uncertainty of surrogate $g$ at an input point $x$, e.g., the mean and standard deviation of a Gaussian process. $\Phi$ and $\phi$ are the cumulative distribution function and probability distribution function of a standard normal random variable. The motivation behind consulting a probabilistic-based infill criterion in EGO is to balance the exploration and exploitation of an optimization process such that the algorithm is unlikely to be trapped by local optima. However, with the assistance of multiple surrogate models, each portraying and capturing distinct landscapes of the objective function, a feasible alternative infill criterion, in comparison with the probabilistic-based ones used in single surrogate-assisted EGO, is the prediction value (PV) of the most promising surrogate model. It can be inferred that the prediction value of a surrogate becomes informative if the promising model being optimized changes drastically iteration-wise since the back-end optimizer can constantly explore different landscapes of the objective function. Rehbach et al.~\cite{rehbach_expected_2020} experimentally discovered that exclusively using prediction values of a predetermined surrogate can outperform using expected improvement in handling higher dimensional (e.g., 10-dimensional) continuous problems. A similar experiment is conducted that compares the performance of SAMA-DiEGO with two infill criteria, i.e., prediction value (PV) and expected improvement (EI), obtained on numerous discrete problems to further discuss the necessity of using a probabilistic-based infill criterion in multiple surrogate-assisted discrete optimizations.

\noindent \textbf{Back-end Optimizer:} In Bayesian optimization, it is expected that the back-end optimizers can efficiently locate the optima of the infill criterion. When handling discrete problems, the non-differentiable search spaces preclude the usage of well-established gradient-based optimizers (e.g., the BFGS algorithm) but favor the usage of robust discrete optimizers (e.g., evolutionary algorithms). Notable non-gradient based optimizers are random local search in~\cite{hutter_sequential_2011}, Genetic Algorithm in~\cite{zaefferer_efficient_2014} and Mixed-Integer Evolutionary Strategy (MIES)~\cite{li_mixed_2013} in~\cite{stein_automatic_2019} and~\cite{horesh_predict_2019}. Following their successes, the SAMA-DiEGO employs MIES for its strong capability of handling both categorical and ordinal variables, which are the two most commonly seen types of discrete variables~\cite{garrido-merchan_dealing_2020}.

\noindent \textbf{Model Selection Mechanism:}
Multiple online model management strategies are discussed in Section~\ref{sec:related_work_A}. Our intention of using multiple surrogates is to allow the optimizer to explore different approximations of the search space provided by distinct (types of) models without requiring additional calls to the objective function. It can be anticipated that auto-tuning strategies (e.g.,~\cite{gorissen_evolutionary_2009}) become cumbersome since they have to cope with a considerable number of hyperparameters at a time. Moreover, the lack of existing studies on model-agnostic uncertainty quantification for discrete surrogate models also precludes the use of ensembles of models (e.g.,~\cite{hanse_optimally_2022}). Additionally, strategies to simultaneously use all surrogate models to find and evaluate promising solutions (e.g., MSEGO~\cite{viana_efficient_2013}) tend to waste expensive function evaluations if some surrogates constantly perform poorly. Therefore, an online model selection technique as proposed in~\cite{bagheri_online_2016} is incorporated in SAMA-DiEGO. Note that in our case, the sliding window mechanism is left out. The sliding window smooths the performance of surrogates across a few iterations and therefore leads to consistency in determining the best surrogate. However, this consistency in model selection contradicts the purpose of applying multiple surrogates in SAMA-DiEGO, and hence the mechanism is excluded.

\begin{algorithm}[tb]
\caption{SAMA-DiEGO. \\
\textbf{Input:} Objective functions $f(\mathbf{x})$, decision parameter space $\mathbf{S}$, number of initial samples $K$, maximum evaluation budget $N_{max}$, model configurations $M$ ($\upvarphi$), number of available parallelisms $P$, infill criterion $I$, time limit $T$. \\
\textbf{Output:} Best found solution $X_{best}$ and its corresponding objective value $y_{best}$.}
\label{alg:SAMA-DiEGO}
\SetAlgoLined
\DontPrintSemicolon
  \SetKwFunction{FMain}{Main}
  \SetKwProg{Pn}{Function}{:}{}
  \Pn{
  \FMain{$f$, $\mathbf{S}$, $K$, $N_{max}$, $M$, $I$, $P$, $T$}}
  {
  $\mathbf{X} \gets \{\mathbf{x}_1, \cdots, \mathbf{x}_{K}$\} \Comment*[f]{Generate initial design, $\mathbf{x_i} \in \mathbf{S}$} 
  
  $\mathbf{Y} \gets f(\mathbf{X})$  \Comment*[r]{Obtain objective scores, $\mathbf{Y} \in \mathbb{R}^{K}$}

$\Tilde{FF}, M^{*} \gets$ \textsc{Verification} ($M$, $\mathbf{X}$, $\mathbf{Y}$, $P$, $T$) \Comment*[f]{Verification of surrogates}

$\hat{m} \gets \textsc{GetFirstOf} (M^{*})$ \Comment*[r]{Get the configurations of the best model after verification}

$n\gets K$

\While{$n \leq N_{max}$} {
$\Tilde{FF}\gets []$ \Comment*[r]{Initialize a pool of surrogates}

$\hat{\mathbf{Y}}\gets$ \textsc{Standardize}$(\mathbf{Y})$ 

\For(\Comment*[f]{For each model configuration}){$\upvarphi \in M^{*}$}{
$\hat{f}_{\mathbf{X}, \mathbf{\hat{Y}}} \gets$ \textsc{Fit}($\upvarphi, \mathbf{X}, \mathbf{\hat{Y}}$) \Comment*[r]{Create and fit models}

$\Tilde{FF} \gets [\Tilde{FF}\ \hat{f}_{\mathbf{X}, \mathbf{\hat{Y}}}]$ \Comment*[r]{Add models into pool}}

$\bar{\mathbf{y}}\gets \textsc{Mean}(\mathbf{Y})$ \Comment*[r]{Get the mean of Objective Values}

$\sigma\gets \textsc{Std}(\mathbf{Y})$ \Comment*[r]{Get the standard deviation}

$\mathbf{y}_{best}\gets$ \textsc{UpdateBest} ($\mathbf{Y}$) \Comment*[r]{Update the best-so-far}
$\mathbf{y}^{*}_{best}\gets (\mathbf{y}_{best}-\bar{\mathbf{y}})\ /\ \sigma$ \Comment*[r]{Standardization}

$\hat{f}^{*}_{\mathbf{X}, \mathbf{\hat{Y}}} \gets \textsc{GetModelByConfig} (\Tilde{FF}, \hat{m})$ \Comment*[f]{Get the best model by configuration}

$\mathbf{IC}\gets I$($\hat{f}^{*}_{\mathbf{X}, \mathbf{\hat{Y}}}$, $\mathbf{y}^{*}_{best}$) \Comment*[f]{Configure the infill criterion}

$\mathbf{x}_{n}\gets$ \textsc{Optimize}($\mathbf{IC}$, $\hat{f}^{*}_{\mathbf{X}, \mathbf{\hat{Y}}}$, $P$, $\mathbf{X}$) \Comment*[f]{Find best new solution w.r.t $\mathbf{IC}$}

$\mathbf{y}_{n}\gets$ $f(\mathbf{x}_{n})$ \Comment*[r]{Evaluate the new solution}

$\mathbf{y}^{*}_{n}\gets {(\mathbf{y}_{n}-\bar{\mathbf{y}}})\ /\ \sigma$ \Comment*[r]{Standardize the new sample}

$\hat{f}_{n}\gets$\textsc{RankModel}($\Tilde{FF}, \mathbf{x}_{n}, \mathbf{y}^{*}_{n}$) \Comment*[f]{Rank and select the best model}

$\hat{m}\gets$\textsc{GetConfigOfModel}$(\hat{f}_{n})$

$\mathbf{X}\gets$ $\left[ \mathbf{X} \; \mathbf{x}_{n} \right]$ \Comment*[r]{Add the new solution, $\mathbf{X} \in \mathbb{R}^{d \times (n+1)}$}
$\mathbf{Y}\gets$ $\left[ \mathbf{Y} \; \mathbf{y}_{n} \right]$ \Comment*[r]{Add the objective value, $\mathbf{Y} \in \mathbb{R}^{n+1}$}

$n\gets n + 1$
}
$\mathbf{x}_{best},\ \mathbf{y}_{best}\gets$ $\textsc{FindBest}$ ($\mathbf{X},\ \mathbf{Y}$) \Comment*[r]{The best w.r.t. $\mathbf{Y}$}
}

\KwRet $(\mathbf{x}_{best},\ \mathbf{y}_{best})$
\end{algorithm}

\subsection{A Walk-through of SAMA-DiEGO}
After reasoning about the choices made for SAMA-DiEGO. A step-by-step introduction to SAMA-DiEGO is provided, starting by explaining the initialization stage, which consists of the following two steps:
\begin{enumerate}[wide, labelwidth=1pt, labelindent=0pt, label=\textbf{\arabic*}), itemsep=0.5em]
    \item Determine the initial design and evaluate these designs on the objective function (lines 2 and 3 in Alg.~\ref{alg:SAMA-DiEGO}). In the absence of historical data, heuristic sampling is performed to obtain the initial samples. Specifically, Latin Hypercube Sampling (LHS) was chosen as suggested for discrete problems in~\cite{horesh_predict_2019} (see section~\ref{sec: setup alg})
    \item With the initial samples, the algorithm verifies all available candidate surrogate models to rule out relatively poorly performing ones according to Feasibility and Performance (line 4 in Alg.~\ref{alg:SAMA-DiEGO}):
    \begin{description}
        \item[Feasibility:]
        Stork et al.~\cite{stork_open_2020} identified a new challenge as the feasibility of existing models. Therefore, with the initial data samples the surrogates are validated and the ineffectual ones are dropped. (e.g., indefinite Kriging kernels). 
        \item[Performance:] 
        The applied ranking criterion $C$ is a train-test split $R^2$ validation. Models are fitted with training samples (70\% of total) and ranked regarding their obtained $R^2$ scores on the validation samples (the 'remaining' 30\%).
        Meanwhile, all candidate models that require more than $T$ seconds to converge are disqualified. As a result, the top $P$ surrogates are chosen.
    \end{description}
    
\end{enumerate}

The verification phase yields a pool of promising surrogate models maintained according to their rankings in the first stage, i.e., the best-performing model and its configuration (parameters) is the first of the $\Tilde{FF}$ and $M^{*}$. These models are subsequently utilized in the second stage (the optimization), which is enclosed in a while-loop in lines 7-27 of Alg.~\ref{alg:SAMA-DiEGO}. Being fundamentally based on the Efficient Global Optimization algorithm, the optimization loop consists of five steps:
\begin{enumerate}[wide, labelindent=0pt, itemsep=0.5em, label=\textbf{\arabic*})]
    \item Standardize all obtained objective values $\mathbf{Y}$ as follows:
    \begin{equation}
    \hat{\mathbf{Y}} = \frac{\mathbf{Y} - \bar{\mathbf{y}}}{\sigma},
    \end{equation}
    where $\bar{\mathbf{y}}$ and $\sigma$ are the mean and standard deviation of $\mathbf{Y}$ respectively, and $\hat{\mathbf{Y}}$ is the output (line 9).
    \item Create and fit candidate surrogate models with standardized data samples (i.e.,$\mathbf{X}$ and $\hat{\mathbf{Y}}$) (lines 10-13). This process is done in parallel, where $P$ surrogate models are concurrently fitted with the same data in $P$ processes.
    \item Set up the acquisition function based on the previously selected surrogate model and the standardized best-so-far objective score (lines 14-19). The determined surrogate model is the best-performing one in the previous iteration (see step 5) or the best in the verification stage in the case of the first iteration of the optimization loop.
\item Search and evaluate the next evaluation point (line 20). Through optimizing the selected infill criterion from step 3, i.e., maximizing EI or finding the optimal PV, the SAMA-DiEGO locates the next promising solution (line 20) and obtains its real objective value regarding the objective function (line 21).
    
\item Rank surrogate models and select the best. The procedure for determining the best surrogate model $\hat{f}_n$ based on the $n$th evaluated data sample in the $n$th iteration is as follows:
\begin{equation}
     \hat{f}_n = \argmin _{\hat{f}_{\mathbf{X}, \mathbf{Y}}\in \Tilde{FF}} \lvert \hat{f}_{\mathbf{X}, \mathbf{Y}}(\mathbf{x}_n) - \mathbf{y}_{n}^{*} \rvert,
\end{equation}
where $\Tilde{FF}$ is the pool of trained surrogate models, $\mathbf{x}_n$ and $\mathbf{y}_n^{*}$ are the $n$th evaluated solution and its real objective value respectively.
\end{enumerate}

The above optimization loop is repeated until the maximum allowed number of evaluations on the objective function (budget) is reached. Lastly, the algorithm returns the best-found evaluated solution $X_{best}$ and its real objective value, $y_{best}$.

\section{Experimental Setup}
\label{sec:exp setup}
Experiments have been carried out to validate the newly proposed SAMA-DiEGO algorithms on discrete benchmark problems, and before diving into the results, setup of these experiments is first discussed in this section.

\subsection{The Used Surrogate Modeling Techniques}
A total of thirty-one distinct surrogate models that naturally support or can be realistically extended to support discrete variables are used in this study. These models belong to four surrogate classes:\\

\textbf{Radial Basis Function network} (RBFn): 
The augmented RBFn applied in~\cite{de_winter_samo-cobra_2021} are chosen for its unrestricted uncertainty quantification which enables us to compute EI on an arbitrary RBFn.
Let $\varphi (d)$ be a radial basis function defined on a Euclidean-distance-based measurement $d$. Nine radial basis functions, including \textit{linear} with $\varphi(d) = d$, \textit{polyharmonic 4} with $\varphi(d) = d^{4}\cdot \log (d)$, \textit{polyharmonic 5} with $\varphi(d) = d^{5}$ and other six functions that have been used in~\cite{de_winter_samo-cobra_2021}, were experimented in this research.\\

\textbf{Kriging interpolation (Kriging)}: A universal Kriging model (a.k.a. Gaussian process regression) is a linear combination of a basis function added to a random process with zero mean and a covariance function~\cite{bouhlel_python_2019}. A convenient implementation of Kriging models provided in the SMT toolbox~\cite{bouhlel_python_2019} is applied in our experiments. This implementation realized a novel transformation mechanism that was proposed in~\cite{garrido-merchan_dealing_2020}, which practically allows Kriging to model discrete problems. Consequently, five correlation functions are selected, namely \textit{Ornstein–Uhlenbeck process, Squared Gaussian correlation, Mat{\'e}rn 3/2, Mat{\'e}rn 5/2, and Gower autocorrelation}, from the SMT toolbox. Each covariance function is then combined with one of the three basis functions (\textit{constant, linear, and quadratic}) to form a Kriging model, and hence 15 in total.\\

\textbf{Random Forest regression (RF)}:
Random forest is an ensemble of decision trees, and it uses the mean of the predictions of all trees to be its prediction. Our utilized RF followed the implementation of van Stein et al.~\cite{stein_automatic_2019}, and it can quantify the uncertainty of a prediction by calculating the variance of the predictions of ensembled decision trees. Hence, this permits us to compute the expected improvement in RF.\\

\textbf{Support-Vector Machines regression (SVM)}:
SVM aims to determine a set of hyperplanes that maximizes the linear separability of data. Given data that is non-linearly separable, the SVM uses kernel functions (e.g., RBF and polynomial) to map data to higher dimensionality. The well-established $\epsilon-$SVM implementation is chose of scikit-learn~\cite{pedregosa_scikit-learn_2011} (v0.24.2) in this research. Six kernels, namely \textit{linear, rbf and sigmoid}, and three \textit{poly}nomial kernels with 2, 3, and 5 degrees, are considered in this research. To the best of our knowledge, studies on quantifying the uncertainty for SVMs do not exist, hence the backend optimizer searches for the next evaluation point exclusively regarding the predictions of SVM regression.

To summarize, the surrogate pool of SAMA-DiEGO contains nine RBFn, fifteen Kriging,  one RF regression, and six SVM regression models. The hyperparameters of all these thirty-one models are not specifically tuned but set according to settings in original implementations or literature.

\subsection{Benchmarked Problems}
Discrete problems can be divided into two classes in regards to the type of variables: ordinal (e.g., integer-valued) and combinatorial (e.g., binary)~\cite{garrido-merchan_dealing_2020}. With regard to the two classes, two experiments are designed, each representing a type of variable, to benchmark the performance of SAMA-DiEGO against several robust optimization algorithms.
\begin{enumerate}[wide, labelindent=0pt, itemsep=0.5em, label=\textbf{\arabic*})]

\item The first problem set is sampled from the Black Box Optimization Benchmark (BBOB) \textit{minimization} problems~\cite{hansen_real-parameter_2009}, where the discretization method proposed in~\cite{tusar_mixed-integer_2019} was applied to transform the continuous input space into an ordinal one. To be distinguished from combinatorial problems with large dimensionality, the search space is configured to be limited in dimensionality but abundant in choices per variable. Moreover, a preliminary study was carried out to categorize the problems into three groups regarding the fidelity of surrogate models~\cite{qi_huang_2022_6684443}. Consequently, fifteen problems are randomly and evenly selected from the three groups and considered in dimensionality 15 with 101 ordinal choices (0 to 100) per variable:
\begin{description}[itemsep=0.2em]
    \item[1 ]: F6 Attractive Sector Function  
    \item[2 ]: F7 Step Ellipsoidal Function
    \item[3 ]: F9 Rosenbrock Function, rotated 
    \item[4 ]: F8 Rosenbrock Function
    \item[5 ]: F11 Discus Function
    \item[6 ]: F13 Sharp Ridge Function
    \item[7 ]: F15 Rastrigin Function
    \item[8 ]: F16 Weierstrass Function
    \item[9 ]: F18 Schaffers F7 Function, moderately ill-conditioned
    \item[10]: F19 Composite Griewank-Rosenbrock Function F8F2
    \item[11]: F20 Schwefel Function
    \item[12]: F23 Katsuura Function
    \item[13]: F21 Gallagher's Gaussian 101-me Peaks Function
    \item[14]: F22 Gallagher's Gaussian 21-hi Peaks Function
    \item[15]: F24 Lunacek bi-Rastrigin Function 
\end{description}
\item The second set is selected from the IOH Pseudo-Boolean Optimization (PBO) benchmark~\cite{doerr_benchmarking_2020}, which features 23 well-organized binary-encoded combinatorial \textit{maximization} problems. Five problems were chosen from the benchmark following the recommendations of~\cite{horesh_predict_2019}:
\begin{description}[itemsep=0em]
    \item[LABS]: low auto-correlation binary sequences problem
    \item[Ising1D Ring] Ising model defined on one-dimensional ring
    \item[Ising2D Torus] Ising model defined on a two-dimensional torus
    \item[MIVS]: maximum independent vertex set problem
    \item[NQP]: N-queens problem
\end{description}
The dimensionalities of problems are determined also in compliance with the settings of~\cite{horesh_predict_2019}, which are \{25, 64, 100\}. The global optima of the five functions are known regardless of dimensionality.
\end{enumerate}

\subsection{Benchmarked Algorithms}
\label{sec: setup alg}
\begin{table*}[htb]
\centering
\resizebox{1\textwidth}{!}{%
\begin{threeparttable}
\caption{\textnormal{The benchmarked algorithms.}}
\begin{tabular}{c|c|llll|c}
\hline
Algorithm & Surrogate & \multicolumn{4}{c|}{Specification} & Experiment \\ \hline
MIES & None & \multicolumn{4}{l|}{Mixed-Integer Evolutionary Strategy~\cite{li_mixed_2013}.} & First \\ \hline
Adaptive TPE & TPE & \multicolumn{4}{l|}{Adaptive Tree-structured Parzen Estimator approach~\cite{bergstra_making_2013}} & First \\ \hline
SMAC & RF & \multicolumn{4}{l|}{Sequential Model Algorithm Configuration~\cite{hutter_sequential_2011}} & First \\ \hline
MIP-EGO & RF & \multicolumn{4}{l|}{Mixed-Integer Parallel Efficient Global Optimization~\cite{stein_automatic_2019}} & Both \\ \hline
P3-GOMEA & None & \multicolumn{4}{l|}{Parameterless Population Pyramid Gene-Pool Optimal Mixing Algorithm~\cite{dushatskiy_novel_2021}} & First \\ \hline
$(1+10)$ EA & None & \multicolumn{4}{l|}{Five $(1+\lambda)$ evolutionary algorithms with a population size of 10.~\cite{doerr_benchmarking_2020}} & Second \\ \hline
(1+(25, 25)) GA & None & \multicolumn{4}{l|}{A self-adjusting genetic algorithm with a population size of 25.~\cite{doerr_benchmarking_2020}} & Second \\ \hline
CatES & None & \multicolumn{4}{l|}{Categorical evolutionary strategy~\cite{horesh_predict_2019}} & Second \\ \hline
SVM-CatES & SVM & \multicolumn{4}{l|}{Support Vector Machines regression assisted CatES~\cite{horesh_predict_2019}} & Second \\ \hline
\multirow{4}{*}{SAMA-DiEGO} & \multirow{4}{*}{\begin{tabular}[c]{@{}c@{}} 9 RBF, \\6 SVMs, \\ 15 Kriging, \\1 RF\end{tabular}} & \multicolumn{1}{c|}{\multirow{2}{*}{Type}} & \multicolumn{3}{c|}{Hyperchoices} & \multirow{2}{*}{-} \\
 &  & \multicolumn{1}{c|}{} & \multicolumn{1}{c}{Infill Criterion ($I$)} & \multicolumn{1}{c}{$P$} & \multicolumn{1}{c|}{$T$} &  \\ \cline{3-7} 
 &  & \multicolumn{1}{c|}{PV} & \multicolumn{1}{l|}{Prediction Value} & \multicolumn{1}{c|}{\multirow{2}{*}{7 parallelisms}} & \multicolumn{1}{c|}{\multirow{2}{*}{30 seconds}} & Both \\ \cline{3-4} \cline{7-7} 
 &  & \multicolumn{1}{c|}{EI} & \multicolumn{1}{l|}{Expected Improvement} & \multicolumn{1}{l|}{} &  & Both \\ \hline
\end{tabular}%
\smallskip
\scriptsize
\begin{tablenotes}
\item Surrogate: the type(s) of surrogate model used by the algorithm;
\item Experiment: on which problem set(s) do(es) the algorithm(s) got benchmarked, \textit{first} for IOH problems, \textit{second} for discretized BBOB, and \textit{both} for both problem sets.
\end{tablenotes}
\label{tab: algorithms}
\end{threeparttable}
}
\end{table*}
The benchmarked algorithms are listed in Table~\ref{tab: algorithms}. The first algorithm, MIES, is the backend optimizer of both MIP-EGO and the proposed SAMA-DiEGO. It serves as the very baseline algorithm for the ordinal BBOB problems. MIP-EGO~\cite{stein_automatic_2019}, adaptive TPE~\cite{bergstra_making_2013} and SMAC~\cite{hutter_sequential_2011} are three efficient and widely-used single-surrogate-assisted EGO algorithms that can solve both discrete and continuous problems. P3-GOMEA is a state-of-the-art version of the gene-pool optimal mixing algorithm that can tackle discrete problems~\cite{dushatskiy_novel_2021}. Four types of algorithms were selected specifically for IOH pseudo-boolean problems, where the $(1+\lambda)$ EA and (1+($\lambda, \lambda$)) GA are existing robust non-surrogate-assisted algorithms suggested by~\cite{doerr_benchmarking_2020}. Specifically, each of the five experimented EAs features a distinct strategy to control mutation rates. CatES and SVM-CatES were experimented by Horesh et al.~\cite{horesh_predict_2019} on the same five IOH pseudo-boolean problems (with the same settings), therefore our results can directly be compared to the results of~\cite{horesh_predict_2019}. Lastly, two proposed SAMA-DiEGO algorithms were experimented and compared to analyze the necessity of using probabilistic-based infill criterion in this context. The (hyper)parameters of all algorithms, except for SAMA-DiEGO, were set in compliance with their publically available implementations and referential literature. Moreover, the two hyperparameters of SAMA-DiEGO (i.e., $P$ and $T$ in Alg.~\ref{alg:SAMA-DiEGO}) were exclusively determined with respect to our available computational resources (i.e., octa-processes per usage). 

Apart from individual hyperparameters of each algorithm, a parameter not to be neglected is the size of initial samples for all surrogate-assisted algorithms. For the IOH pseudo-boolean problems, the initial sampling size was set to the dimensionality of the problem, which was primarily bounded by the applied Latin hypercube sampling strategy of SAMA-DiEGO. And following the settings of~\cite{rehbach_expected_2020}, the size of initial sampling is set to 10\% of the total budget (function calls to the objective function) for the experiments on discretized BBOB problems.

With regard to the aforementioned setups, a fixed-budget experiment is proposed to analyze and compare the performance of the algorithms. All algorithms are allowed to use up to 500 function calls for a given optimization problem. Moreover, each algorithm was independently executed 11 times on a given problem to (maximally) stabilize the performance under our computational resources.
\section{Results}
\label{sec:results}
\subsection{The Ordinal BBOB Problems}
\begin{table}
\centering
\resizebox{0.9\textwidth}{!}{%
\begin{threeparttable}
\captionsetup{font=normalsize}
\caption{\textnormal{Benchmark results on discretized (ordinal) BBOB minimization problems led by \textbf{SAMA-DiEGO}. For each test problem, {\color[HTML]{6433FC}blue} highlights the algorithm that achieved the lowest average best-found fitness value across 11 independent runs. Whereas, the other results highlighted in {\color[HTML]{32CB00}green} are incomparable to the best solution according to a Wilcoxon rank sum test with $\alpha=0.05$.}}
\label{tab:BBOB_SAMA}
\begin{tabular}{ll|llll}
\hline
\multicolumn{1}{c}{} & \multicolumn{1}{c|}{} & \multicolumn{4}{c}{Dimension = 15} \\ \cline{3-6} 
\multicolumn{1}{c}{} & \multicolumn{1}{c|}{} & \multicolumn{4}{c}{Fitness Value} \\
\multicolumn{1}{c}{\multirow{-3}{*}{Function}} & \multicolumn{1}{c|}{\multirow{-3}{*}{Algorithm}} & \multicolumn{1}{c}{Best} & \multicolumn{1}{c}{Mean} & \multicolumn{1}{c}{Median} & \multicolumn{1}{c}{Std} \\ \hline
 & MI-ES & 172869.15 & 407604.29 & 428103.05 & 209158.25 \\
 & MIP-EGO & 21227.12 & 61558.02 & 53880.90 & 13787.24 \\
 & Adaptive TPE & 387.80 & 6884.71 & 7191.29 & 5945.59 \\
 & {\color[HTML]{32CB00} SMAC} & {\color[HTML]{32CB00} 72.05} & {\color[HTML]{32CB00} 99.15} & {\color[HTML]{32CB00} 99.16} & {\color[HTML]{32CB00} 16.20} \\
 & P3-GOMEA & 148.02 & 3884.69 & 12891.50 & 18872.99 \\
 & {\color[HTML]{6433FC} SAMA-DiEGO-PV} & {\color[HTML]{6433FC} 76.35} & {\color[HTML]{6433FC} 93.44} & {\color[HTML]{6433FC} 94.86} & {\color[HTML]{6433FC} 17.81} \\
\multirow{-7}{*}{F6 Attractive Sector} & SAMA-DiEGO-EI & 233.04 & 11822.81 & 14705.90 & 10733.29 \\ \hline
 & MI-ES & 361.10 & 682.51 & 765.20 & 314.34 \\
 & MIP-EGO & 223.94 & 269.91 & 276.73 & 32.11 \\
 & Adaptive TPE & 139.31 & 164.78 & 177.11 & 25.68 \\
 & SMAC & 111.35 & 126.16 & 128.40 & 11.47 \\
 & P3-GOMEA & 156.84 & 209.53 & 206.48 & 28.50 \\
 & {\color[HTML]{6433FC} SAMA-DiEGO-PV} & {\color[HTML]{6433FC} 104.07} & {\color[HTML]{6433FC} 114.74} & {\color[HTML]{6433FC} 117.48} & {\color[HTML]{6433FC} 11.39} \\
\multirow{-7}{*}{F7 Step Ellipsoidal} & SAMA-DiEGO-EI & 146.07 & 180.11 & 188.11 & 31.78 \\ \hline
 & MI-ES & 43955.62 & 104023.45 & 111845.50 & 46027.14 \\
 & MIP-EGO & 14296.34 & 17477.50 & 17779.66 & 2520.72 \\
 & Adaptive TPE & 1302.25 & 4796.49 & 5614.59 & 2990.26 \\
 & SMAC & 565.94 & 2225.38 & 4038.31 & 4345.35 \\
 & P3-GOMEA & 42008.57 & 59620.57 & 69886.26 & 25249.70 \\
 & {\color[HTML]{6433FC} SAMA-DiEGO-PV} & {\color[HTML]{6433FC} 283.74} & {\color[HTML]{6433FC} 606.00} & {\color[HTML]{6433FC} 609.30} & {\color[HTML]{6433FC} 191.79} \\
\multirow{-7}{*}{F8 Rosenbrock Original} & SAMA-DiEGO-EI & 1073.39 & 5357.91 & 4928.82 & 2720.72 \\ \hline
 & MI-ES & 44756.84 & 73232.62 & 79370.41 & 31158.36 \\
 & MIP-EGO & 5123.26 & 14679.66 & 13981.47 & 5139.36 \\
 & Adaptive TPE & 1401.03 & 2181.28 & 2284.56 & 753.30 \\
 & SMAC & 500.69 & 2478.33 & 2293.00 & 1502.34 \\
 & P3-GOMEA & 579.97 & 1547.67 & 2241.56 & 1597.13 \\
 & {\color[HTML]{6433FC} SAMA-DiEGO-PV} & {\color[HTML]{6433FC} 214.83} & {\color[HTML]{6433FC} 256.11} & {\color[HTML]{6433FC} 316.88} & {\color[HTML]{6433FC} 159.97} \\
\multirow{-7}{*}{F9 Rosenbrock Rotated} & SAMA-DiEGO-EI & 214.83 & 426.35 & 581.77 & 328.22 \\ \hline
 & MI-ES & 206.25 & 279.45 & 207713.44 & 452104.85 \\
 & MIP-EGO & 166.11 & 233.11 & 337.58 & 276.67 \\
 & Adaptive TPE & 125.13 & 217.01 & 229.29 & 78.24 \\
 & {\color[HTML]{32CB00} SMAC} & {\color[HTML]{32CB00} 135.17} & {\color[HTML]{32CB00} 187.86} & {\color[HTML]{32CB00} 199.90} & {\color[HTML]{32CB00} 41.14} \\
 & P3-GOMEA & 160.71 & 205.91 & 207.61 & 27.70 \\
 & {\color[HTML]{6433FC} SAMA-DiEGO-PV} & {\color[HTML]{6433FC} 152.51} & {\color[HTML]{6433FC} 179.92} & {\color[HTML]{6433FC} 185.90} & {\color[HTML]{6433FC} 19.26} \\
\multirow{-7}{*}{F11 Discus} & SAMA-DiEGO-EI & 168.88 & 215.24 & 217.75 & 30.85 \\ \hline
 & MI-ES & -82.41 & -70.90 & -70.03 & 11.00 \\
 & MIP-EGO & -92.40 & -90.98 & -90.82 & 1.12 \\
 & Adaptive TPE & -96.42 & -95.68 & -95.44 & 0.79 \\
 & SMAC & -97.05 & -94.95 & -94.90 & 1.03 \\
 & P3-GOMEA & -96.13 & -94.03 & -94.05 & 1.20 \\
 & {\color[HTML]{6433FC} SAMA-DiEGO-PV} & {\color[HTML]{6433FC} -102.30} & {\color[HTML]{6433FC} -96.38} & {\color[HTML]{6433FC} -96.91} & {\color[HTML]{6433FC} 2.11} \\
\multirow{-7}{*}{\begin{tabular}[c]{@{}l@{}}F19 Composite \\ Griewank-Rosenbrock\end{tabular}} & SAMA-DiEGO-EI & -96.90 & -94.76 & -95.00 & 0.87 \\ \hline
 & MI-ES & 25393.04 & 72963.97 & 76338.17 & 33981.88 \\
 & MIP-EGO & 1580.41 & 5264.21 & 5734.52 & 2624.57 \\
 & Adaptive TPE & -514.11 & -128.67 & 66.48 & 474.78 \\
 & SMAC & -543.55 & 298.68 & 1000.94 & 1143.17 \\
 & P3-GOMEA & -542.92 & -201.39 & 564.72 & 1157.05 \\
 & {\color[HTML]{6433FC} SAMA-DiEGO-PV} & {\color[HTML]{6433FC} -544.35} & {\color[HTML]{6433FC} -543.55} & {\color[HTML]{6433FC} -543.59} & {\color[HTML]{6433FC} 0.40} \\
\multirow{-7}{*}{F20 Schwefel} & SAMA-DiEGO-EI & -535.31 & -366.33 & -193.45 & 416.77 \\ \hline
\end{tabular}%
\end{threeparttable}
}
\end{table}

\begin{table}
\centering
\resizebox{0.9\textwidth}{!}{%
\begin{threeparttable}
\captionsetup{font=normalsize}
\caption{\textnormal{Benchmark results on discretized (ordinal) BBOB minimization problems. For each test problem, {\color[HTML]{6433FC}blue} highlights the algorithm that achieved the lowest average best-found fitness value across 11 independent runs. Whereas, the other results highlighted in {\color[HTML]{32CB00}green} are incomparable to the best solution according to a Wilcoxon rank sum test with $\alpha=0.05$.}}
\label{tab:BBOB_tie}
\begin{tabular}{ll|llll}
\hline
\multicolumn{1}{c}{} & \multicolumn{1}{c|}{} & \multicolumn{4}{c}{Dimension = 15} \\ \cline{3-6} 
\multicolumn{1}{c}{} & \multicolumn{1}{c|}{} & \multicolumn{4}{c}{Fitness Value} \\
\multicolumn{1}{c}{\multirow{-3}{*}{Function}} & \multicolumn{1}{c|}{\multirow{-3}{*}{Algorithm}} & \multicolumn{1}{c}{Best} & \multicolumn{1}{c}{Mean} & \multicolumn{1}{c}{Median} & \multicolumn{1}{c}{Std} \\ \hline
 & MI-ES & 15.54 & 29.97 & 31.97 & 11.10 \\
 & MIP-EGO & 0.80 & 17.16 & 17.39 & 7.09 \\
 & {\color[HTML]{6433FC} Adaptive TPE} & {\color[HTML]{6433FC} -9.57} & {\color[HTML]{6433FC} 0.11} & {\color[HTML]{6433FC} -0.56} & {\color[HTML]{6433FC} 3.12} \\
 & {\color[HTML]{32CB00} SMAC} & {\color[HTML]{32CB00} -9.13} & {\color[HTML]{32CB00} 0.31} & {\color[HTML]{32CB00} 0.48} & {\color[HTML]{32CB00} 6.99} \\
 & P3-GOMEA & 0.65 & 6.62 & 5.86 & 3.06 \\
 & SAMA-DiEGO-PV & -3.17 & 1.19 & 1.04 & 2.93 \\
\multirow{-7}{*}{\begin{tabular}[c]{@{}l@{}}F18 Schaffers F7 \\ (moderately ill-conditioned)\end{tabular}} & {\color[HTML]{32CB00} SAMA-DiEGO-EI} & {\color[HTML]{32CB00} -4.82} & {\color[HTML]{32CB00} 0.44} & {\color[HTML]{32CB00} 1.02} & {\color[HTML]{32CB00} 4.57} \\ \hline
 & MI-ES & 9.92 & 11.13 & 11.21 & 0.69 \\
 & {\color[HTML]{32CB00} MIP-EGO} & {\color[HTML]{32CB00} 8.99} & {\color[HTML]{32CB00} 10.65} & {\color[HTML]{32CB00} 10.34} & {\color[HTML]{32CB00} 0.75} \\
 & {\color[HTML]{32CB00} Adaptive TPE} & {\color[HTML]{32CB00} 9.50} & {\color[HTML]{32CB00} 10.63} & {\color[HTML]{32CB00} 10.52} & {\color[HTML]{32CB00} 0.76} \\
 & {\color[HTML]{32CB00} SMAC} & {\color[HTML]{32CB00} 9.20} & {\color[HTML]{32CB00} 10.55} & {\color[HTML]{32CB00} 10.65} & {\color[HTML]{32CB00} 0.58} \\
 & {\color[HTML]{32CB00} P3-GOMEA} & {\color[HTML]{32CB00} 9.03} & {\color[HTML]{32CB00} 10.30} & {\color[HTML]{32CB00} 10.27} & {\color[HTML]{32CB00} 0.53} \\
 & SAMA-DiEGO-PV & 9.85 & 10.81 & 10.86 & 0.68 \\
\multirow{-7}{*}{F23 Katsuura} & {\color[HTML]{6433FC} SAMA-DiEGO-EI} & {\color[HTML]{6433FC} 9.64} & {\color[HTML]{6433FC} 10.16} & {\color[HTML]{6433FC} 10.30} & {\color[HTML]{6433FC} 0.60} \\ \hline
 & MI-ES & 358.86 & 437.38 & 456.27 & 71.83 \\
 & MIP-EGO & 282.88 & 320.01 & 319.88 & 15.99 \\
 & {\color[HTML]{32CB00} Adaptive TPE} & {\color[HTML]{32CB00} 253.99} & {\color[HTML]{32CB00} 280.45} & {\color[HTML]{32CB00} 276.14} & {\color[HTML]{32CB00} 10.94} \\
 & {\color[HTML]{32CB00} SMAC} & {\color[HTML]{32CB00} 246.68} & {\color[HTML]{32CB00} 279.30} & {\color[HTML]{32CB00} 288.88} & {\color[HTML]{32CB00} 33.66} \\
 & P3-GOMEA & 276.85 & 295.09 & 294.50 & 11.66 \\
 & {\color[HTML]{32CB00} SAMA-DiEGO-PV} & {\color[HTML]{32CB00} 245.91} & {\color[HTML]{32CB00} 303.27} & {\color[HTML]{32CB00} 297.76} & {\color[HTML]{32CB00} 31.59} \\
\multirow{-7}{*}{F24 Lunacek bi-Rastrigin} & {\color[HTML]{6433FC} SAMA-DiEGO-EI} & {\color[HTML]{6433FC} 229.96} & {\color[HTML]{6433FC} 276.68} & {\color[HTML]{6433FC} 276.03} & {\color[HTML]{6433FC} 21.78} \\ \hline
\end{tabular}%
\end{threeparttable}
}
\end{table}

\begin{table}
\centering
\resizebox{0.9\textwidth}{!}{%
\begin{threeparttable}
\captionsetup{font=normalsize}
\caption{\textnormal{Benchmark results on discretized (ordinal) BBOB problems led by \textbf{SMAC}. For each test problem, {\color[HTML]{6433FC}blue} highlights the algorithm that achieved the lowest average best-found fitness value across 11 independent runs. Whereas, the other results highlighted in {\color[HTML]{32CB00}green} are incomparable to the best solution according to a Wilcoxon rank sum test with $\alpha=0.05$.}}
\label{tab:BBOB_others}
\begin{tabular}{ll|llll}
\hline
\multicolumn{1}{c}{} & \multicolumn{1}{c|}{} & \multicolumn{4}{c}{Dimension = 15} \\ \cline{3-6} 
\multicolumn{1}{c}{} & \multicolumn{1}{c|}{} & \multicolumn{4}{c}{Fitness Value} \\
\multicolumn{1}{c}{\multirow{-3}{*}{Function}} & \multicolumn{1}{c|}{\multirow{-3}{*}{Algorithm}} & \multicolumn{1}{c}{Best} & \multicolumn{1}{c}{Mean} & \multicolumn{1}{c}{Median} & \multicolumn{1}{c}{Std} \\ \hline
 & MI-ES & 1805.62 & 2177.03 & 2186.14 & 225.16 \\
 & MIP-EGO & 943.29 & 1009.77 & 1042.65 & 87.45 \\
 & Adaptive TPE & 668.11 & 758.38 & 762.45 & 84.39 \\
 & {\color[HTML]{6433FC} SMAC} & {\color[HTML]{6433FC} 258.09} & {\color[HTML]{6433FC} 417.76} & {\color[HTML]{6433FC} 412.22} & {\color[HTML]{6433FC} 88.60} \\
 & P3-GOMEA & 656.29 & 856.28 & 855.72 & 145.99 \\
 & {\color[HTML]{32CB00} SAMA-DiEGO-PV} & {\color[HTML]{32CB00} 268.56} & {\color[HTML]{32CB00} 514.22} & {\color[HTML]{32CB00} 492.19} & {\color[HTML]{32CB00} 122.41} \\
\multirow{-7}{*}{F13 Sharp Ridge} & SAMA-DiEGO-EI & 625.11 & 827.61 & 863.90 & 146.71 \\ \hline
 & MI-ES & 1337.55 & 1544.87 & 1572.05 & 223.00 \\
 & MIP-EGO & 1231.29 & 1341.62 & 1327.01 & 51.03 \\
 & Adaptive TPE & 1173.00 & 1198.76 & 1201.05 & 20.33 \\
 & {\color[HTML]{6433FC} SMAC} & {\color[HTML]{6433FC} 1098.96} & {\color[HTML]{6433FC} 1170.15} & {\color[HTML]{6433FC} 1163.20} & {\color[HTML]{6433FC} 30.73} \\
 & P3-GOMEA & 1177.40 & 1217.94 & 1226.36 & 36.73 \\
 & {\color[HTML]{32CB00} SAMA-DiEGO-PV} & {\color[HTML]{32CB00} 1123.21} & {\color[HTML]{32CB00} 1172.55} & {\color[HTML]{32CB00} 1167.24} & {\color[HTML]{32CB00} 23.65} \\
\multirow{-7}{*}{\begin{tabular}[c]{@{}l@{}}F15 Rastrigin \\ Moderately Ill-conditioned\end{tabular}} & SAMA-DiEGO-EI & 1172.35 & 1207.25 & 1217.00 & 34.37 \\ \hline
 & MI-ES & 84.24 & 100.79 & 96.79 & 7.81 \\
 & MIP-EGO & 89.66 & 96.05 & 96.33 & 3.50 \\
 & Adaptive TPE & 87.63 & 98.08 & 96.99 & 3.23 \\
 & {\color[HTML]{6433FC} SMAC} & {\color[HTML]{6433FC} 85.14} & {\color[HTML]{6433FC} 92.29} & {\color[HTML]{6433FC} 91.49} & {\color[HTML]{6433FC} 3.10} \\
 & P3-GOMEA & 91.99 & 96.09 & 96.96 & 3.46 \\
 & {\color[HTML]{32CB00} SAMA-DiEGO-PV} & {\color[HTML]{32CB00} 85.10} & {\color[HTML]{32CB00} 93.22} & {\color[HTML]{32CB00} 93.25} & {\color[HTML]{32CB00} 5.70} \\
\multirow{-7}{*}{F16 Weierstrass} & SAMA-DiEGO-EI & 90.34 & 96.64 & 96.23 & 3.63 \\ \hline
 & MI-ES & 97.56 & 112.49 & 111.21 & 6.81 \\
 & MIP-EGO & 75.47 & 91.13 & 91.16 & 9.26 \\
 & Adaptive TPE & 53.32 & 60.82 & 62.80 & 8.61 \\
 & {\color[HTML]{6433FC} SMAC} & {\color[HTML]{6433FC} 42.45} & {\color[HTML]{6433FC} 51.12} & {\color[HTML]{6433FC} 49.88} & {\color[HTML]{6433FC} 4.96} \\
 & P3-GOMEA & 51.45 & 72.57 & 73.36 & 15.09 \\
 & SAMA-DiEGO-PV & 49.80 & 71.38 & 74.09 & 12.93 \\
\multirow{-7}{*}{\begin{tabular}[c]{@{}l@{}}F21 Gallagher's \\ Gaussian 101-me Peaks\end{tabular}} & {\color[HTML]{32CB00} SAMA-DiEGO-EI} & {\color[HTML]{32CB00} 46.81} & {\color[HTML]{32CB00} 51.16} & {\color[HTML]{32CB00} 56.28} & {\color[HTML]{32CB00} 7.76} \\ \hline
 & MI-ES & -925.42 & -917.58 & -918.79 & 3.07 \\
 & MIP-EGO & -959.34 & -937.17 & -942.87 & 10.24 \\
 & Adaptive TPE & -977.57 & -948.56 & -952.38 & 12.52 \\
 & {\color[HTML]{6433FC} SMAC} & {\color[HTML]{6433FC} -997.29} & {\color[HTML]{6433FC} -979.23} & {\color[HTML]{6433FC} -982.75} & {\color[HTML]{6433FC} 10.64} \\
 & P3-GOMEA & -980.78 & -943.83 & -952.91 & 14.21 \\
 & SAMA-DiEGO-PV & -963.68 & -939.61 & -943.27 & 13.19 \\
\multirow{-7}{*}{\begin{tabular}[c]{@{}l@{}}F22 Gallagher's \\ Gaussian 21-hi Peaks\end{tabular}} & SAMA-DiEGO-EI & -978.12 & -971.07 & -966.68 & 11.30 \\ \hline
\end{tabular}%

\end{threeparttable}
}
\end{table}

\begin{table*}
\caption{\textnormal{The experimental results obtained on five PBO problems defined on 25, 64, and 100 dimensions. The results are highlighted in blue if they are significantly better according to a Wilcoxon rank sum test with $\alpha =0.05$ over eleven independent runs.}}
\label{tab: PBO result}
\resizebox{\textwidth}{!}{%
\begin{threeparttable}
\begin{tabular}{ll|cccc|cccc|cccc}
\hline
\multicolumn{1}{c}{} & \multicolumn{1}{c|}{} & \multicolumn{4}{c|}{Dimension = 25} & \multicolumn{4}{c|}{Dimension = 64} & \multicolumn{4}{c}{Dimension = 100} \\ \cline{3-14} 
\multicolumn{1}{c}{} & \multicolumn{1}{c|}{} & \multicolumn{3}{c|}{Objective Value} & \multicolumn{1}{l|}{Budgets} & \multicolumn{3}{c|}{Objective Value} & \multicolumn{1}{l|}{Budgets} & \multicolumn{3}{c|}{Objective Value} & \multicolumn{1}{l}{Budgets} \\
\multicolumn{1}{c}{\multirow{-3}{*}{Functions}} & \multicolumn{1}{c|}{\multirow{-3}{*}{Algorithms}} & \multicolumn{1}{l}{Mean} & \multicolumn{1}{l}{Std} & \multicolumn{1}{l|}{Best} & \multicolumn{1}{l|}{Mean$^\dag$} & \multicolumn{1}{l}{Mean} & \multicolumn{1}{l}{Std} & \multicolumn{1}{l|}{Best} & \multicolumn{1}{l|}{Mean$^\dag$} & \multicolumn{1}{l}{Mean} & \multicolumn{1}{l}{Std} & \multicolumn{1}{l|}{Best} & \multicolumn{1}{l}{Mean$^\dag$} \\ \hline
Ising1D & CatES & 21.68 & N/A & \multicolumn{1}{c|}{25.00} & N/A & 48.24 & N/A & \multicolumn{1}{c|}{54.00} & N/A & 71.16 & N/A & \multicolumn{1}{c|}{76.00} & N/A \\
 & SVM-CatES & 23.40 & N/A & \multicolumn{1}{c|}{25.00} & N/A & 53.56 & N/A & \multicolumn{1}{c|}{56.00} & N/A & 77.44 & N/A & \multicolumn{1}{c|}{84.00} & N/A \\
 & MIP-EGO & 21.50 & 0.89 & \multicolumn{1}{c|}{23.00} & 500 & 45.09 & 1.31 & \multicolumn{1}{c|}{48.00} & 500 & 65.46 & 0.89 & \multicolumn{1}{c|}{66.00} & 500 \\
 & (1+10) EA* & 21.73 & 1.54 & \multicolumn{1}{c|}{25.00} & 485 & 52.73 & 1.55 & \multicolumn{1}{c|}{56.00} & 500 & 81.09 & 3.94 & \multicolumn{1}{c|}{86.00} & 500 \\
 & (1+(25, 25)) GA & 21.73 & 0.96 & \multicolumn{1}{c|}{23.00} & 500 & 51.64 & 1.43 & \multicolumn{1}{c|}{54.00} & 500 & 76.73 & 2.74 & \multicolumn{1}{c|}{82.00} & 500 \\
 & SAMA-DiEGO-PV & {\color[HTML]{6433FC} 25.00} & {\color[HTML]{6433FC} 0.00} & \multicolumn{1}{c|}{{\color[HTML]{6433FC} 25.00}} & 259 & {\color[HTML]{6434FC} 64.00} & {\color[HTML]{6434FC} 0.00} & \multicolumn{1}{c|}{{\color[HTML]{6434FC} 64.00}} & 99 & {\color[HTML]{6434FC} 100.00} & {\color[HTML]{6434FC} 0.00} & \multicolumn{1}{c|}{{\color[HTML]{6434FC} 100.00}} & 125 \\
 & SAMA-DiEGO-EI & {\color[HTML]{6433FC} 25.00} & {\color[HTML]{6433FC} 0.00} & \multicolumn{1}{c|}{{\color[HTML]{6433FC} 25.00}} & 291 & {\color[HTML]{6434FC} 64.00} & {\color[HTML]{6434FC} 0.00} & \multicolumn{1}{c|}{{\color[HTML]{6434FC} 64.00}} & 66 & {\color[HTML]{6434FC} 100.00} & {\color[HTML]{6434FC} 0.00} & \multicolumn{1}{c|}{{\color[HTML]{6434FC} 100.00}} & 106 \\ \hline
Ising2D & CatES & 42.04 & N/A & \multicolumn{1}{c|}{50.00} & N/A & 90.50 & N/A & \multicolumn{1}{c|}{98.00} & N/A & 133.44 & N/A & \multicolumn{1}{c|}{152.00} & N/A \\
 & SVM-CatES & 49.92 & N/A & \multicolumn{1}{c|}{50.00} & N/A & 101.64 & N/A & \multicolumn{1}{c|}{116.00} & N/A & 144.44 & N/A & \multicolumn{1}{c|}{154.00} & N/A \\
 & MIP-EGO & 39.27 & 1.07 & \multicolumn{1}{c|}{44.00} & 500 & 83.28 & 1.77 & \multicolumn{1}{c|}{86.00} & 500 & 122.91 & 2.88 & \multicolumn{1}{c|}{126.00} & 500 \\
 & (1+10) EA* & 45.23 & 4.69 & \multicolumn{1}{c|}{50.00} & 407 & 99.28 & 5.00 & \multicolumn{1}{c|}{106.00} & 500 & 153.46 & 4.98 & \multicolumn{1}{c|}{164.00} & 500 \\
 & (1+(25, 25)) GA & 44.36 & 4.07 & \multicolumn{1}{c|}{50.00} & 485 & 96.37 & 3.89 & \multicolumn{1}{c|}{102.00} & 500 & 142.37 & 2.93 & \multicolumn{1}{c|}{146.00} & 500 \\
 & SAMA-DiEGO-PV & {\color[HTML]{6434FC} 50.00} & {\color[HTML]{6434FC} 0.00} & \multicolumn{1}{c|}{{\color[HTML]{6434FC} 50.00}} & 150 & {\color[HTML]{6434FC} 128.00} & {\color[HTML]{6434FC} 0.00} & \multicolumn{1}{c|}{{\color[HTML]{6434FC} 128.00}} & 85 & {\color[HTML]{6434FC} 200.00} & {\color[HTML]{6434FC} 0.00} & \multicolumn{1}{c|}{{\color[HTML]{6434FC} 200.00}} & 123 \\
 & SAMA-DiEGO-EI & {\color[HTML]{6434FC} 50.00} & {\color[HTML]{6434FC} 0.00} & \multicolumn{1}{c|}{{\color[HTML]{6434FC} 50.00}} & 177 & {\color[HTML]{6434FC} 128.00} & {\color[HTML]{6434FC} 0.00} & \multicolumn{1}{c|}{{\color[HTML]{6434FC} 128.00}} & 68 & {\color[HTML]{6434FC} 200.00} & {\color[HTML]{6434FC} 0.00} & \multicolumn{1}{c|}{{\color[HTML]{6434FC} 200.00}} & 110 \\ \hline
NQP & CatES & 4.16 & N/A & \multicolumn{1}{c|}{5.00} & N/A & -36.96 & N/A & \multicolumn{1}{c|}{-3.00} & N/A & -264.30 & N/A & \multicolumn{1}{c|}{24.00} & N/A \\
 & SVM-CatES & 4.56 & N/A & \multicolumn{1}{c|}{5.00} & N/A & 2.64 & N/A & \multicolumn{1}{c|}{6.00} & N/A & -15.20 & N/A & \multicolumn{1}{c|}{26.00} & N/A \\
 & MIP-EGO & -0.73 & 2.67 & \multicolumn{1}{c|}{3.00} & 500 & -247.55 & 29.94 & \multicolumn{1}{c|}{-193.00} & 500 & -753.3 & 53.80 & \multicolumn{1}{c|}{-641.00} & 500 \\
 & (1+10) EA* & 4.36 & 0.64 & \multicolumn{1}{c|}{5.00} & 392 & 3.55 & 5.25 & \multicolumn{1}{c|}{8.00} & 500 & -46.60 & 32.80 & \multicolumn{1}{c|}{7.00} & 500 \\
 & (1+(25, 25)) GA & 3.81 & 0.93 & \multicolumn{1}{c|}{5.00} & 451 & -59.55 & 22.92 & \multicolumn{1}{c|}{-25.00} & 500 & -288.60 & 73.70 & \multicolumn{1}{c|}{-186.0} & 500 \\
 & SAMA-DiEGO-PV & {\color[HTML]{6434FC} 5.00} & {\color[HTML]{6434FC} 0.00} & \multicolumn{1}{c|}{{\color[HTML]{6434FC} 5.00}} & 195 & 3.36 & 0.48 & \multicolumn{1}{c|}{4.00} & 500 & {\color[HTML]{6433FC} 6.00} & {\color[HTML]{6433FC} 1.00} & \multicolumn{1}{c|}{{\color[HTML]{6433FC} 7.00}} & 500 \\
 & SAMA-DiEGO-EI & {\color[HTML]{6434FC} 5.00} & {\color[HTML]{6434FC} 0.00} & \multicolumn{1}{c|}{{\color[HTML]{6434FC} 5.00}} & 260 & {\color[HTML]{6434FC} 6.00} & {\color[HTML]{6434FC} 0.43} & \multicolumn{1}{c|}{{\color[HTML]{6434FC} 7.00}} & 500 & {\color[HTML]{6433FC} 6.30} & {\color[HTML]{6433FC} 0.60} & \multicolumn{1}{c|}{{\color[HTML]{6433FC} 7.00}} & 500 \\ \hline
LABS & CatES & 3.93 & N/A & \multicolumn{1}{c|}{6.01} & N/A & 2.58 & N/A & \multicolumn{1}{c|}{3.58} & N/A & 2.11 & N/A & \multicolumn{1}{c|}{2.73} & N/A \\
 & SVM-CatES & 4.31 & N/A & \multicolumn{1}{c|}{6.51} & N/A & 2.76 & N/A & \multicolumn{1}{c|}{4.03} & N/A & 2.24 & N/A & \multicolumn{1}{c|}{2.94} & N/A \\
 & MIP-EGO & 3.26 & 0.45 & \multicolumn{1}{c|}{4.34} & 500 & 2.17 & 0.20 & \multicolumn{1}{c|}{2.69} & 500 & 1.77 & 0.08 & \multicolumn{1}{c|}{1.91} & 500 \\
 & (1+10) EA* & 4.16 & 0.75 & \multicolumn{1}{c|}{4.88} & 500 & 3.31 & 0.32 & \multicolumn{1}{c|}{3.97} & 500 & {\color[HTML]{6433FC} 3.10} & {\color[HTML]{6433FC} 0.27} & \multicolumn{1}{c|}{{\color[HTML]{6433FC} 3.55}} & 500 \\
 & (1+(25, 25)) GA & 4.01 & 0.94 & \multicolumn{1}{c|}{6.51} & 500 & 2.53 & 0.21 & \multicolumn{1}{c|}{2.83} & 500 & 2.19 & 0.17 & \multicolumn{1}{c|}{2.48} & 500 \\
 & SAMA-DiEGO-PV & {\color[HTML]{6434FC} 4.80} & {\color[HTML]{6434FC} 0.78} & \multicolumn{1}{c|}{{\color[HTML]{6434FC} 6.51}} & 500 & 2.71 & 0.21 & \multicolumn{1}{c|}{3.03} & 500 & 2.77 & 0.31 & \multicolumn{1}{c|}{3.32} & 500 \\
 & SAMA-DiEGO-EI & 4.10 & 0.62 & \multicolumn{1}{c|}{5.58} & 500 & {\color[HTML]{6433FC} 3.69} & {\color[HTML]{6433FC} 0.21} & \multicolumn{1}{c|}{{\color[HTML]{6433FC} 4.23}} & 500 & {\color[HTML]{6433FC} 3.17} & {\color[HTML]{6433FC} 0.34} & \multicolumn{1}{c|}{{\color[HTML]{6433FC} 3.74}} & 500 \\ \hline
MIVS & CatES & 10.84 & N/A & \multicolumn{1}{c|}{12.00} & N/A & -1.00 & N/A & \multicolumn{1}{c|}{24.00} & N/A & -308.60 & N/A & \multicolumn{1}{c|}{-99.00} & N/A \\
 & SVM-CatES & 11.66 & N/A & \multicolumn{1}{c|}{12.00} & N/A & 17.44 & N/A & \multicolumn{1}{c|}{23.00} & N/A & -63.18 & N/A & \multicolumn{1}{c|}{-3.00} & N/A \\
 & MIP-EGO & 8.82 & 0.83 & \multicolumn{1}{c|}{10.00} & 500 & -370.36 & 127.11 & \multicolumn{1}{c|}{-116.00} & 500 & -1657.18 & 263.19 & \multicolumn{1}{c|}{-1069.00} & 500 \\
 & (1+10) EA* & 11.45 & 0.89 & \multicolumn{1}{c|}{12.00} & 271 & {\color[HTML]{6434FC} 26.00} & {\color[HTML]{6434FC} 1.54} & \multicolumn{1}{c|}{{\color[HTML]{6434FC} 29.00}} & 500 & -21.73 & 77.64 & \multicolumn{1}{c|}{36.00} & 500 \\
 & (1+(25, 25)) GA & 10.63 & 1.30 & \multicolumn{1}{c|}{12.00} & 431 & -17.18 & 49.56 & \multicolumn{1}{c|}{21.00} & 500 & -353.36 & 251.72 & \multicolumn{1}{c|}{23.00} & 500 \\
 & SAMA-DiEGO-PV & {\color[HTML]{6434FC} 12.0} & {\color[HTML]{6434FC} 0.00} & \multicolumn{1}{c|}{{\color[HTML]{6434FC} 12.00}} & 276 & 11.45 & 0.99 & \multicolumn{1}{c|}{13.00} & 500 & 23.45 & 6.43 & \multicolumn{1}{c|}{36.00} & 500 \\
 & SAMA-DiEGO-EI & {\color[HTML]{6434FC} 12.0} & {\color[HTML]{6434FC} 0.00} & \multicolumn{1}{c|}{{\color[HTML]{6434FC} 12.00}} & 299 & 22.73 & 2.18 & \multicolumn{1}{c|}{26.00} & 500 & {\color[HTML]{6433FC} 28.73} & {\color[HTML]{6433FC} 4.05} & \multicolumn{1}{c|}{{\color[HTML]{6433FC} 34.00}} & 500 \\ \hline
\end{tabular}
\begin{tablenotes}
\scriptsize
\item[$^\dag$] Each \textit{Budgets Mean} column records the average number of function evaluations over 11 runs used by all algorithms to find the global optimum (if found, otherwise 500) of the problem respectively. Such results for ES and SVM-CatES are not available in case they are not provided in~\cite{horesh_predict_2019}.
\item[$^\ast$] Instead of showing all results obtained by five (1 + 10) EA algorithms, the result of the best-performed (1 + 10) EA according to the significance test is shown per case.
\end{tablenotes}

\end{threeparttable}
}
\end{table*}

The benchmark results of algorithms on the discretized BBOB \textit{minimization} problems are shown in Table~\ref{tab:BBOB_SAMA}, Table~\ref{tab:BBOB_tie}, and Table~\ref{tab:BBOB_others}. Additionally, an analysis on the convergence profiles of this experiment can be found in~\cite{qi_huang_2022_6684443}.

The results in Table~\ref{tab:BBOB_SAMA} are the cases where SAMA-DiEGO with prediction value is generally the best algorithm. Chasing after SAMA-DiEGO-PV, SMAC becomes a competitive runner-up on two test problems (i.e., F6, F11). Next to Table~\ref{tab:BBOB_SAMA}, first Table~\ref{tab:BBOB_tie} is presented where multiple algorithms give comparable results. SAMA-DiEGO with expected improvement becomes twice the best algorithms regarding the best mean objective value, and adaptive TPE leads on F18. Lastly, Table~\ref{tab:BBOB_others} shows cases where SMAC tops the results, and SAMA-DiEGO arrives second on four problems.

Through relating the performance of algorithms to properties of problems as mentioned in~\cite{hansen_real-parameter_2009}, it can be observed that our proposed SAMA-DiEGO is a robust algorithm for single-modal problems but shows a slight deficiency for multi-modal problems in comparison with SMAC. These observations are reasonable since the rapid switch of surrogate models in SAMA-DiEGO helps the algorithm converge on problems with clear global structures, but inevitably distracts the search when handling problems with a considerable number of local structures spread over the whole search space (e.g., F21 and F22). Furthermore, it can also be discovered that prediction value is generally the better infill criterion for SAMA-DiEGO, but expected improvement should be favored if the problems are multi-modal across the entire search space. Lastly, by comparing the results of MIES (non-surrogate), MIPEGO (single-surrogate), and SAMA-DiEGO (multi-surrogate), it can be concluded that using a surrogate model indeed facilitates the optimization of ordinal-spaced problems. And moreover, our proposed strategy, which considers multi-surrogate models, can benefit the search process even more.

\subsection{The Pseudo-Boolean Problems}
The results of the second experiment are presented in Table~\ref{tab: PBO result}. The best-performing algorithms are highlighted in blue if they outperform others in a Wilcoxon rank sum test ($\alpha=0.05$) over 11 independent runs. According to Table~\ref{tab: PBO result}, SAMA-DiEGO performs best on fourteen of the fifteen test cases, with the exception of the 64-dimensional MIVS problems, where the (1+10) EA performed better. Moreover, as it is indicated in \textbf{Budgets Mean} columns of the result table, both SAMA-DiEGOs can find the global optimum of the two Ising problems across three dimensionalities and are also capable of solving NQP and MIVS on 25 dimensions within 500 fitness evaluations. Unexpectedly, the single surrogate assisted algorithms MIP-EGO performed worse than the (1+10) evolutionary algorithms in fourteen of the fifteen cases. Furthermore, because SAMA-DiEGO, SVM-CatES, and MIP-EGO all use MI-ES as a back-end solver, the differences in performance of the three algorithms can be attributed to their respective surrogate models, substantiating the benefit of using our proposed model management strategy. Also, SAMA-DiEGO with EI did better on 64 and 100 dimensions than SAMA-DiEGO with PV. Therefore, it is reasonable to consider that the expected improvement is a booster for SAMA-DiEGO to solve higher dimensional combinatorial problems in comparison with its sibling.

\section{Conclusions and Future works}
\label{sec:conclusion}
In this paper, a novel algorithm, Self-Adaptive Multiple-surrogate Assisted Efficient Global Optimization (SAMA-DiEGO), is proposed. Following the iterative optimization process of Bayesian optimization, the SAMA-DiEGO additionally features a two-stage online model selection mechanism, which iteratively chooses the most suitable surrogate from a pool of candidate models to serve as the cheaper substitute for the discrete objective function. Experiments were carried out to compare SAMA-DiEGO with existing robust optimization algorithms on fifteen binary-encoded combinatorial test cases and fifteen ordinal-encoded problems. The results indicated that SAMA-DiEGO is a robust solver for combinatorial problems, and ordinal problems. Moreover, the performance of SAMA-DiEGO with expected improvement as an infill criterion was compared with that of exclusively using the prediction value of the promising surrogate. In conclusion, prediction value is an efficient and promising infill criterion for multi-surrogate assisted EGO algorithms in general, while expected improvement is better for solving high-dimensional combinatorial problems and highly multi-modal ordinal problems.

Lastly, several foreseeable directions following this research are briefly discussed. Firstly, 
it would be beneficial to track and compare the running time of algorithms in future benchmarking. Secondly, given the fact that all components of SAMA-DiEGO essentially work for both continuous and discrete variables, it would be a rewarding attempt to further investigate the benefit of using SAMA-DiEGO in mixed-integer cases. Other possible improvements are utilizing novel infill criterion (e.g. moment generating function of improvement~\cite{wang_new_2017}) to better tackle multi-modal problems, and employing hybrid dimensionality-reduction methods (e.g., PCA for BO~\cite{raponi_high_2020}) to efficiently handle high-dimensional combinatorial problems. 

\bibliographystyle{plain}  
\bibliography{references}

\end{document}